\pdfoutput=1

\documentclass[11pt]{article}

\usepackage{acl}

\usepackage{times}
\usepackage{latexsym}

\usepackage[T1]{fontenc}

\usepackage[utf8]{inputenc}

\usepackage{microtype}

\usepackage{inconsolata}

\usepackage{xspace}
\usepackage{amssymb}
\usepackage{amsfonts}
\usepackage{subfigure}
\usepackage{todonotes}
\usepackage{pifont}
\usepackage{mathtools}
\usepackage{environ}
\usepackage{multirow}
\usepackage{booktabs}
\usepackage{mdframed}

\newcommand{\model}{DetermLR\xspace}
\DeclareMathOperator*{\argmax}{arg\,max}

\title{DetermLR: Augmenting LLM-based Logical Reasoning \\from Indeterminacy to Determinacy}

\author{Hongda Sun$^{13*}$ \hspace{2mm} Weikai Xu$^{23*}$\hspace{2mm} Wei Liu$^{3}$\hspace{2mm} Jian Luan$^{3}$\hspace{2mm} Bin Wang$^{3}$\hspace{2mm}\\ {\bf Shuo Shang$^{2\dagger}$\hspace{2mm} Ji-Rong Wen$^{1}$\hspace{2mm} Rui Yan$^{1\dagger}$} \\
$^{1}$Gaoling School of Artificial Intelligence, Renmin University of China\\
$^{2}$University of Electronic Science and Technology of China, 
$^{3}$XiaoMi AI Lab\\
\texttt{\{sunhongda98,jrwen,ruiyan\}@ruc.edu.cn, {xuwk266,jedi.shang\}@gmail.com}
}
\\
\texttt{\{liuwei40,luanjian,wangbin11\}@xiaomi.com}
}

\begin{document}
\maketitle
\begin{abstract}
Recent advances in large language models (LLMs) have revolutionized the landscape of reasoning tasks. To enhance the capabilities of LLMs to emulate human reasoning, prior studies have focused on modeling reasoning steps using various thought structures like chains, trees, or graphs. However, LLM-based reasoning still encounters the following challenges: (1) Limited adaptability of preset structures to diverse tasks; (2) Insufficient precision in exploiting known conditions to derive new ones; and (3) Inadequate consideration of historical reasoning experiences for subsequent reasoning steps. To this end, we propose DetermLR, a novel perspective that rethinks the reasoning process as an evolution from indeterminacy to determinacy. First, we categorize known conditions into two types: determinate and indeterminate premises This provides an oveall direction for the reasoning process and guides LLMs in converting indeterminate data into progressively determinate insights. Subsequently, we leverage quantitative measurements to prioritize more relevant premises to explore new insights. Furthermore, we automate the storage and extraction of available premises and reasoning paths with reasoning memory, preserving historical reasoning details for subsequent reasoning steps. Comprehensive experimental results demonstrate that DetermLR surpasses all baselines on various logical reasoning benchmarks: LogiQA, ProofWriter, FOLIO, PrOntoQA, and LogicalDeduction. Compared to previous multi-step reasoning methods, DetermLR achieves higher accuracy with fewer reasoning steps, highlighting its superior efficiency and effectiveness in solving logical reasoning tasks.
\end{abstract}

\section{Introduction}
The emergence of large language models (LLMs) has instigated a transformative wave within the realm of artificial intelligence~\citep{zhao2023survey}.
The series models of GPT~\citep{gpt3,instructgpt,gpt4} and PaLM~\citep{palm,palm2} have exhibited remarkable proficiency in natural language reasoning, contributing to the advancement of research and applications of cognitive intelligence~\citep{huang2022survey}.
However, even the current state-of-the-art (SOTA) LLMs still grapple with a key limitation: the lack of human-like advanced reasoning skills to rationally analyze known conditions and draw conclusions~\citep{gpt4cant,singh2023mind}.
This leaves a substantial gap between LLM-based reasoning and the cognitive process of human reasoning.

To alleviate this limitation, existing studies employ enhanced prompt engineering techniques to guide LLMs in eliciting intermediate thinking steps to ensure reliable conclusions~\citep{zhou2022least,khot2022decomposed,wei2022chain,zerocot}.
Building upon this foundation, recent works have focused on introducing more intricate reasoning structures, such as multiple chains~\citep{wang2022self}, trees~\citep{tot} or graphs~\citep{lei2023boosting,besta2023graph}, to tackle increasingly complex reasoning tasks.
However, LLM-based reasoning continues to encounter three challenges: 
(1) Limited adaptability of preset structures to diverse tasks: Since the task complexity cannot be solely inferred from the problem context, relying on a certain preset structure to solve a variety of reasoning problems may create deficiencies in reasoning effectiveness or efficiency~\citep{tot,lei2023boosting}. 
This approach contrasts with human problem-solving techniques, which are not dependent on preset reasoning structures. 
Ideally, the reasoning structure should be the result of manual review after solving the problem.
(2) Insufficient precision in exploiting known conditions to derive new ones:
The literature on human cognitive reasoning provides valuable insights and emphasizes the importance of integrating available information for informed decision-making~\citep{schaeken1999deductive,evans2002logic,baron2023thinking}.
This motivates cumulative reasoning (CR)~\citep{cr}, which uses LLMs to iteratively generate new propositions based on available premises.
However, CR still cannot approach the human thought process, as it relies on the random combination of existing premises without a well-defined criterion.
(3) Inadequate consideration of historical reasoning experiences for future reasoning: Previous works~\citep{wei2022chain,tot} often overlook historical reasoning details, resulting in the lack of necessary information for subsequent phases.

To address these challenges and augment LLMs to grasp more human-like advanced reasoning skills, we need to consider three key factors:
(1) Refine the formulation of the essence of the reasoning process;
(2) Prioritize relevant premises for efficiently exploring new information;
(3) Memorize historical reasoning details to guide the direction of the subsequent reasoning steps.

To this end, we propose \model, a novel reasoning framework to align LLM-based reasoning more closely with human thinking.
First, we formulate the logical reasoning process as an evolution from indeterminacy to determinacy.
Since premises exhibit varying descriptions and associations with the target, we initiate the reasoning process with premise identification to finely categorize premises into two distinct types: determinate and indeterminate. Determinate premises are defined as simple statements, which can definitively contribute to conclusion derivation.
Conversely, indeterminate premises may contain complex rules governing the relationships among multiple propositions. 
Regardless of the problem complexity, the reasoning process consistently involves the continuous accumulation of determinate information, steering the conclusion toward greater clarity.

Subsequently, human reasoning typically aims for a ``breakingthrough'' from known conditions to deduce new insights, indicating the necessity to distinguish the priority of premises.
Therefore, we propose quantitative measurements to facilitate premise prioritization, which involves identifying the most relevant premise to the conclusion and screening supplementary premises likely to interact with this primary premise.
This guides LLMs to exclude irrelevant premises and focus on more pertinent information for premise exploration.

Furthermore, we introduce a reasoning memorization module to automate the storage and extraction of available premises and reasoning paths.
In this way, historical reasoning details are preserved in the reasoning memory to update reasoning states, and they are incorporated into future reasoning steps to refer to inherent experiences and avoid repeating similar mistakes.

To verify the capability of LLMs to engage in rigorous logical reasoning,
we conduct extensive experiments on various challenging logical reasoning benchmarks: LogiQA, ProofWriter, FOLIO, ProntoQA, and LogicalDeduction.
The experimental results show that \model achieves the best performance on reasoning accuracy, coupled with superior efficiency of requiring fewer steps than previous multi-step reasoning methods.
Notably, in more intricate tasks like LogiQA, \model exhibits even more pronounced advancements, mirroring human-like reasoning skills to a greater extent.

Our technical contributions to advancing LLM-based reasoning can be summarized as follows:

$\bullet$ We propose a novel framework that formulates the reasoning process as an evolution from indeterminacy to determinacy, aligning LLM-based reasoning more closely with human reasoning.

$\bullet$ We leverage quantitative measurements for premise prioritization and exploration, enabling LLMs to prioritize premises more conducive to exploring new insights and deriving conclusions.

$\bullet$ We introduce a reasoning memorization module to preserve essential historical reasoning details during the iterative reasoning process.

\section{Related Work}

\subsection{Conventional Logical Reasoning}
Many previous works focus on solving logical reasoning tasks using machine reading comprehension~\cite{ouyang2021fact}, adversarial pre-training~\cite{pi2022logigan}, and contrastive learning~\cite{jiao2022merit}.
In the realm of logical reasoning benchmarks, various tasks have been utilized for evaluation~\citep{khot2018scitail,wang2022lsat,bhagavatula2019abductive,welleck2018dialogue,williams2017broad,dagan2005pascal,bowman2015large,wang2018glue,liu2021natural,tian2021diagnosing}.
LogiQA~\citep{liu2020logiqa} involves diverse types of logical reasoning questions collected from the National Civil Servants Examination of China.
Based on Big-Bench~\citep{srivastava2022beyond}, which is used to evaluate multi-aspect abilities of language models, Big-Bench Hard (BBH)~\citep{suzgun2022challenging} focuses on 23 challenging tasks for evaluating LLM-based reasoning.
FOLIO~\citep{han2022folio} is a human-annotated and logically complex datasets for natural language reasoning, equipped with first-order logic (FOL) annotations. ProofWriter~\citep{tafjord2020proofwriter} is another commonly used dataset for deductive logical reasoning.
Among them, we carefully select five benchmarks whose premises are listed directly in the problem context and require no additional processing.


\begin{figure*}
    \centering
    \includegraphics[width=\textwidth]{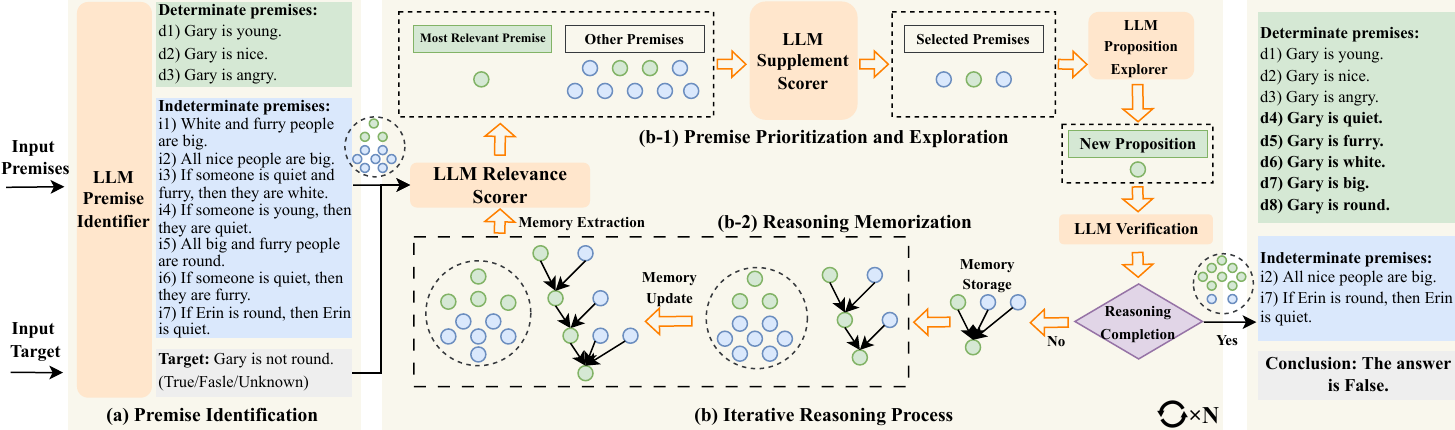}
    \caption{The overview of DetermLR: (a) premise identification; (b) iterative reasoning process: (b-1) premise prioritization and exploration and (b-2) reasoning memorization. Green elements represent determinate premises, and blue elements represent indeterminate premises. The proportion of blue decreases with the accumulation of green during iterative reasoning.}
    \label{fig:model}
\end{figure*}

\subsection{LLM-based Logical Reasoning}
Previous methods mainly enhance reasoning by eliciting intermediate steps like
chain-of-thought (CoT)~\citep{wei2022chain,wang2022self} and least-to-most prompting~\citep{zhou2022least}.
Extending the CoT concept, which follows a left-to-right progression, more recent works present more intricate thought structures to face more complex problems, such as ToT~\citep{tot,hu2023tree} or GoT~\citep{lei2023boosting}.
Selection-inference~\citep{creswell2022selection} refine the reasoning process of CoT by decomposing it into two modules: selection and inference.
Algorithm-of-Thoughts~\citep{sel2023algorithm}  navigate reasoning pathways as in-context examples with merely a few queries.
Cumulative reasoning~\citep{cr} uses higher-order logic rules for exploring new propositions based on given premises.
Current LLM-based reasoning methods still face challenges in emulating human-like reasoning skills.
In response to these challenges, we propose a novel perspective for formulating the reasoning process. This approach prioritizes the exploration of new insights by leveraging more relevant premises and enables iterative reasoning based on key historical reasoning experiences.

\section{DetermLR}
\subsection{Problem Formulation}
The objective of a logical reasoning problem can be regarded as using known premises and logical deduction rules to derive new essential intermediate propositions,  culminating in an eventual target conclusion. 
Suppose a problem provides a set of $N$ premises, denoted as $\mathcal{P} = (p_1, p_2, \cdots, p_N)$, the logical reasoning process can be formulated as:
\begin{equation}
    c = \texttt{Reason}(p_1, p_2, \cdots, p_n),
\end{equation}
where $c$ is the target conclusion of the problem, and the mapping \texttt{Reason} indicates how to use the given premises to derive the conclusion.
In this paper, our focus is on aligning LLM-based reasoning more closely with human reasoning. Therefore, the \texttt{Reason} is generally implemented by instructing LLMs to understand the problem and iteratively provide new insights to solve it.

Building upon the available premises and the target as input, we propose a novel perspective to formulate the process of logical reasoning.
In the following sections, we will introduce in detail three pivotal modules of the proposed method: (1) premise identification (\S \ref{sec:identify}); (2) premise prioritization and exploration (\S \ref{sec:priotrity}); and (3) iterative reasoning memorization (\S \ref{sec:memory}).

 
\subsection{Premise Identification}\label{sec:identify}
As previously discussed, the essence of the reasoning process lies in the mapping \textit{from premises to conclusions}. 
Existing methods~\cite{tot,besta2023graph} that preset the reasoning structure before solving a problem may not always yield an appropriate mapping, as this diverges from the focus of human reasoning.
Instead, the so-called reasoning structure should be formed based on the reviewed reasoning outcomes after problem resolution.
Thus, refining a better formulation for the essence of the reasoning process becomes the first key factor in augmenting LLM-based reasoning.
Regarding the given premises in a logical reasoning problem, it becomes apparent that the "determinacy" of the information supplied by each premise varies significantly: some directly provide pertinent information for deriving the conclusion, while others necessitate combination with other conditions to obtain new insights.
The indeterminacy gradually diminishes with the accumulation of determinate information, bringing the reasoning process closer to the conclusion.
Therefore, we rethink the essence of the reasoning process as \textbf{from indeterminacy to determinacy}.
To be more specific, we develop a premise identification module to emulate the transition from indeterminacy to determinacy.
Input premises are categorized into two distinct types: determinate premises $\mathcal{D}$ and indeterminate premises $\mathcal{I}$.
The identification criterion is dependent upon both the inherent description of the premise and its connection with the input target $c$, as expressed by:
\begin{equation}
    \mathcal{D}, \mathcal{I} = \texttt{Identify}(\mathcal{P}, c).
\end{equation}
In practice, we implement the \texttt{Identify} function through carefully designed instructions for LLMs, and related prompt templates are available in Appendix.
Determinate premises are defined as simple statements that definitively lead to the desired target. These premises state clarified facts and serve as the foundational blocks for reasoning.
In contrast, indeterminate premises encompass propositions not directly related to the target and often contain complex structures reflecting indeterminacy, such as disjunction ($x$ or $y$) and hypothesis (if $x$ then $y$).
An indeterminate premise may be combined with other premises to establish a logical path to evolve into a determinate state.

As shown in Figure~\ref{fig:model}(a), the target revolves around ``Gary'' and ``round'', so simple statements including ``Gary'' are identified as determinate premises ($d_1\cdots d_3$), while the remaining premises are classified as indeterminate ones ($i_1\cdots i_7$).
Building upon this module, LLMs can eliminate the need for preset structures and enhance the clarity of the reasoning process under our new formulation.

\subsection{Premise Prioritization and Exploration}\label{sec:priotrity}
Once the original premises are categorized, how to better uncover the relationships between these premises to explore new insights is the next critical reasoning step.
Prior sampling-based methods cannot distinguish the priority of different premises~\cite{tot,cr}, leading to less skillful reasoning compared to human counterparts.
Therefore, we aim to quantify the relationship between each premise and the target and prioritize premise selection for better exploration.

\paragraph{Premise prioritization with two-stage scoring.}
To improve the opportunity of deriving new insights, we leverage two quantitative measurements to select useful premises for combination.
Firstly, we evaluate the relevance score of each premise $p$ in conjunction with the target $c$. By simulating the overlap of topics and elements within them, varying priorities can be assigned to the premises.
Secondly, we select the most relevant premise $p_*$ from $\mathcal{D}$ as the primary premise, and all other premises are considered candidate supplementary premises to interact with $p_*$.
We then quantify the likelihood of these premises being merged with $p_*$ according to semantic similarity and adherence to logical deduction rules. 
Therefore, supplementary premises $\mathbf{p}_s$ exceeding a given threshold $\theta$ can be obtained.
The two-stage scoring can be formulated as:
\begin{align}
    r_p & = \texttt{relevance}(p, c), \quad p_* \triangleq \argmax_{p \in \mathcal{D}} r_p, \\
    s_{p'} & = \texttt{supplement}(p_*, p'), \nonumber \\
    & \quad \  \mathbf{p}_s \triangleq \{p' \in \mathcal{D}\cup\mathcal{I}\setminus \{p_*\}; s_{p'}\ge\theta \},
\end{align}
where both the \texttt{relevance} and \texttt{supplement} functions are implemented through carefully designed instructions for LLMs. See Appendix for detailed prompt templates.


\paragraph{Premise exploration with three-fold verification.}
Once selected premises for exploration are determined, we employ LLMs to execute the \texttt{explore} function, which considers combining supplementary premises $\mathbf{p}_s$ with the primary premise $p_*$ to generate a new proposition $\widehat{p}$, which can be given by: 
\begin{align}
    \widehat{p} = \texttt{explore}(p_*, \mathbf{p}_s).
\end{align}
Next, the rationality of the newly explored proposition $\widehat{p}$ undergoes rigorous verification, encompassing three-fold critical aspects:
(1) \textit{Logical validity}: We verify whether the deduction of the selected premises to $\widehat{p}$ is valid in terms of logical reasoning rules; (2) \textit{Useful contribution}: We verify whether $\widehat{p}$ is a useful determinate premise that contributes to derive the conclusion. It helps filter out the ``correct nonsense'' that may be logically valid but fail to enhance the conclusion derivation;
(3) \textit{Duplication avoidance}: We verify whether $\widehat{p}$ provides information gain beyond the original premises, avoiding the generation of mere paraphrases of existing premises.
Only propositions that pass all these verification checks will be retained and added to the determinate premise set.
The main steps of premise prioritization and exploration can be formulated as:

{\fontsize{10.4}{0}\selectfont
\begin{equation}
    \mathcal{D} \leftarrow \mathcal{D}\cup\{\widehat{p}\}, \text{ if } \texttt{verify}(\widehat{p}, \{p_*, \mathbf{p}_s\}) = \text{True},
\end{equation}
}
where the \texttt{verify} function is also implemented by the carefully designed instructions for LLMs, and detailed prompt templates are available in Appendix.
Through premise prioritization and exploration, LLMs can effectively prioritize more pertinent premises to explore new insights, improving reasoning effectiveness and efficiency.

\subsection{Reasoning Memorization}\label{sec:memory}
As known conditions dynamically update during the reasoning process, conventional methods often overlook historical reasoning details, resulting in erroneous reasoning directions or stagnant reasoning progress~\citep{tot,cr}. 
In contrast, humans generally record previous reasoning steps and retain both successful and failed attempts in mind to continue reasoning. 
To bridge this cognitive gap, we design a reasoning memorization module to automate the storage and extraction of available premises and evolving reasoning structures.
We initialize the reasoning memory as $\mathcal{M} = \mathcal{D}^{(0)}\cup\mathcal{I}^{(0)}$, only containing input premises before the reasoning process.
Figure~\ref{fig:model}(c) illustrates an iteration of memory storage and extraction, which is elaborated in detail as follows.

\paragraph{Memory storage.} During the $t$-th iteration of premise exploration, our focus of the reasoning details lies on the new proposition $\widehat{p}^{(t)}$ and the reasoning paths $G_{\widehat{p}^{(t)}}$ that connect the original premises $\{p_*^{(t)}, \mathbf{p}_s^{(t)}\}$ to $\widehat{p}^{(t)}$. If $\widehat{p}^{(t)}$ passes all verification checks, we denote the reasoning paths as positive $G_{\widehat{p}^{(t)}}^+$ and 
store both $\widehat{p}^{(t)}$ and $G_{\widehat{p}^{(t)}}^+$
into the reasoning memory.
Otherwise, the reasoning paths will be designated as negative $G_{\widehat{p}^{(t)}}^-$ and also stored into the memory as part of the reasoning experiences. This process can be formulated as:
{\fontsize{9.2}{0}\selectfont
\begin{equation}
\mathcal{M}^{(t)} = \left\{
\begin{aligned}
    \mathcal{M}^{(t-1)}\cup \{\widehat{p}^{(t)}, G_{\widehat{p}^{(t)}}^+\}, & \quad \text{if verify = True} \\
    \mathcal{M}^{(t-1)}\cup \{G_{\widehat{p}^{(t)}}^-\}, & \quad \text{otherwise} \\
\end{aligned}
\right.
\end{equation}
}
\paragraph{Memory extraction.}
When we consider prioritizing premises in the $(t+1)$-th iteration, we extract $t$ previous reasoning details from memory to guide LLMs in drawing upon successful experiences and avoiding repetitive mistakes.
Following each iteration of premise exploration, it is essential to extract current premises and reasoning paths from memory. This extraction can help accurately verify whether the current determinate information is sufficient to draw the target conclusion.
More details about the reasoning memory are available in Appendix.

Overall, the reasoning memory supports both retrospective and prospective reasoning during the iterative process. Retrospectively, it stores historical reasoning details for updating reasoning states. Prospectively, it extracts previous reasoning experiences into future steps, enhancing the accuracy of premise prioritization and exploration.

\begin{table*}
  \centering
  \caption{Comparison results on LogiQA, ProofWriter, FOLIO, and LogicalDeduction. Bold numbers highlight the highest accuracy and the fewest steps among multi-step methods.}
  \resizebox{\textwidth}{!}{
    \begin{tabular}{@{}llcccccccccc@{}}
    \toprule
    \midrule
    \multirow{2}[4]{*}{\textbf{Model}} & \multirow{2}[4]{*}{\textbf{Method}} & \multicolumn{2}{c}{\textbf{LogiQA}} & \multicolumn{2}{c}{\textbf{ProofWriter}} & \multicolumn{2}{c}{\textbf{FOLIO}} &
    \multicolumn{2}{c}{\textbf{PrOntoQA}} & \multicolumn{2}{c}{\textbf{LD}} \\
\cmidrule{3-12}          &       & \textbf{Accuracy} $\uparrow$ & \textbf{Avg. Steps} $\downarrow$ & \textbf{Accuracy} $\uparrow$ & \textbf{Avg. Steps} $\downarrow$ & \textbf{Accuracy} $\uparrow$ & \textbf{Avg. Steps} $\downarrow$ & \textbf{Accuracy} $\uparrow$ & \textbf{Avg. Steps} $\downarrow$ & \textbf{Accuracy} $\uparrow$ & \textbf{Avg. Steps} $\downarrow$ \\
    \midrule
    \multirow{8}[4]{*}{GPT-3.5-turbo} & Standard &   16.76    &   1    & 36.17 &    1   & 49.51 & 1 & 51.80 & 1  &   41.33    & 1 \\
          & CoT   &   22.35    &    1   & 45.00 &     1  & 54.41 &    1   &  84.00 & 1 &  46.00   & 1 \\
          \cmidrule{2-12}
          & CoT-SC &   22.91    &   16    & 48.67 &     16  & 57.34 &   16    & 86.80  & 16 &  50.33    & 16 \\
          & SI    & 24.02 & 15.16 & 50.17 & 18.49 & 57.84 & 14.19 & 88.60  & 13.58 & 51.00 & 17.24 \\
    & LAMBADA & 24.02 & 59.32 & 55.17 & 16.89 & 60.29 & 12.35 & 90.80  & 12.09 & 62.67 & 74.43 \\
          & ToT   & 26.25 &  19.87     & 54.16 &  24.88     & 59.80 & 19.82      & 91.20  & 19.30 & 66.33 & 23.71 \\
          & CR & 31.84 &   18.93    & 59.16 &  18.81     & 59.80 & 18.96      & 92.40 & 16.93 & 71.00 & 18.32 \\
\cmidrule{2-12}          & \textbf{DetermLR} & \textbf{37.99} & \textbf{13.39}       & \textbf{68.83} &  \textbf{16.52}     & \textbf{63.72} &  \textbf{10.37}     & \textbf{93.20} & \textbf{10.74} & \textbf{74.33} & \textbf{13.19} \\
\midrule
    \multirow{8}[4]{*}{GPT-4} & Standard & 31.69 & 1     & 46.83 & 1     & 60.29 & 1     & 77.40  & 1 & 71.33 & 1 \\
          & CoT   & 38.55 & 1     & 67.41 & 1     & 67.65 & 1  & 91.00    & 1    & 73.33 & 1 \\
          \cmidrule{2-12}
          & CoT-SC & 40.43 & 16    & 69.33 & 16    & 68.14 & 16 & 93.40  & 16    & 74.67 & 16 \\
          & SI    & 41.34 & 14.35 & 70.67 & 17.46 & 69.11 & 13.76 & 93.80  & 11.38 & 76.33 & 14.95 \\
    & LAMBADA & 39.11 & 56.24 & 72.00 & 15.04 & 70.10 & 10.85 & 95.60  & 10.56 & 78.00 & 67.32 \\
          & ToT   & 43.02 & 19.87  & 70.33 & 24.57  & 69.12 & 19.12 & 97.60  & 18.91   & 76.83 & 21.83  \\
          & CR & 45.25 & 17.00    & 71.67 & 16.76  & 69.11 & 15.87 & 98.20  & 14.18 & 78.33 & 16.98 \\
\cmidrule{2-12}          & \textbf{DetermLR} & \textbf{54.19} & \textbf{11.74} & \textbf{79.17} & \textbf{14.63} & \textbf{75.49} & \textbf{8.57} & \textbf{98.60} & \textbf{9.78} & \textbf{85.00} & \textbf{12.35} \\
\midrule
    \bottomrule
    \end{tabular}}
  \label{tab:main}
\end{table*}

\section{Experiments}
\subsection{Experimental Setup}
\paragraph{Datasets.}
To verify the capability of LLMs to engage in rigorous logical reasoning based solely on established conditions, without external knowledge, we carefully select five challenging logical reasoning benchmarks: 
(1) \textbf{LogiQA}~\citep{liu2020logiqa} collects the multiple-choice logical problems from National Civil Servants Examination of China.
Since it contains different types of questions, we carefully reviewed its test set and retained 179 high-quality questions whose premises are delineated within the context as a curated collection.
(2) \textbf{ProofWriter}~\citep{tafjord2020proofwriter} is a widely used logical reasoning benchmark. We use the open-world assumption subset where each case requires to be proven true, false or unknown. 
We follow~\citet{pan2023logic} to use the depth-5 subset containing 600 cases for evaluation.
(3) \textbf{FOLIO}~\citep{han2022folio} is a challenging benchmark requiring complex first-order logic reasoning to solve. 
We follow the official data split and choose the validation set containing 204 examples for evaluation.
(4) \textbf{PrOntoQA}~\cite{prontoqa} is similar to ProofWriter for evaluating logical reasoning.
(5) \textbf{LogicalDeduction (LD)} is a challenging task in BigBench~\citep{srivastava2022beyond}.
The problems are mainly about deducing the order of objects from a set of conditions. We use the full test set containing 300 examples for evaluation. 
Logical reasoning examples for each task are available in Appendix.





\paragraph{Baselines.}

To compare our DetermLR with existing LLM-based reasoning methods, we choose the following baselines: 1) \textbf{Standard} prompting directly answers the question based on in-context examples; 2) \textbf{CoT}~\citep{wei2022chain} adopts step-by-step generation of indeterminate rationales before the final answer; 3) \textbf{CoT-SC}~\citep{wang2022self} uses majority voting to aggregate multiple CoTs; 
4) \textbf{SI}~\cite{si} uses selection-inference patterns for iterative reasoning;
5) \textbf{LAMBADA}~\cite{kazemi2022lambada} performs backward chaining for automated reasoning tasks;
6) \textbf{ToT}~\citep{tot} models the reasoning process as a thought search tree; 7) \textbf{CR}~\citep{cr} is recently proposed to generate new propositions based on available premises.

In principle, our proposed framework imposes no restrictions on the type of used LLMs.
Here we uniformly employ the most advanced GPT-4~\citep{gpt4} and GPT-3.5-turbo as the base model to test the upper limit of LLM-based logical reasoning.
Our implementation is based on the Microsoft guidance library~\footnote{\url{https://github.com/guidance-ai/guidance}.}. 
We set the temperature to 0.1 by default and 0.7 for CoT-SC ($n=16$).

\subsection{Main Results}

The results presented in Table~\ref{tab:main} demonstrate that our proposed \model achieves superior reasoning accuracy with fewer steps compared to other multi-step reasoning methods (CR and ToT). While all methods show improvement over GPT-3.5-turbo by approximately 0.1 accuracy, DetermLR consistently outperforms all baselines even on the same base model.
For the most challenging LogiQA, all baselines including CR perform poorly on this task with accuracy below 46.
Since the utilization order of known conditions is crucial to solving the exam problem, baseline methods often fail to grasp the accurate reasoning direction.
\model performs well by prioritizing and memorizing known conditions and reasoning steps, resulting in an accuracy of 54.19 with GPT-4.
Meanwhile, the average number of reasoning steps in \model reaches 11.74, which is more efficient than CoT-SC, ToT and CR in solving real logical reasoning examination problems.
For ProofWriter and FOLIO, \model can generate more accurate propositions for the target than CoT-SC, ToT, and CR.
Also, \model requires fewer reasoning steps to reach the same conclusion, ensuring more efficient premise integration and exploration.
The results in LD shows that compared to all baseline methods, \model can enhance the accuracy of assigning the order of objects, and substantially reduce the number of reasoning iterations.

\begin{table}
  \centering
  \caption{Ablation results: accuracy (first row) and average reasoning steps (second row in parentheses).}
  \resizebox{\linewidth}{!}{
    \begin{tabular}{@{}lcccc@{}}
    \toprule
    \textbf{Method} & \textbf{LogiQA} & \textbf{ProofWriter} & \textbf{FOLIO} & \textbf{LD} \\
    \midrule
    \multirow{2}[1]{*}{\model w/o identify} & 46.15 & 71.50 & 69.61 & 79.00 \\
          & (17.24)  & (16.58)  & (13.70)  & (16.84)  \\
    \multirow{2}[0]{*}{\model w/o priority} & 47.83 & 72.32 & 70.59 & 80.33 \\
          & (18.35)  & (17.21)  & (14.69)  & (17.02)  \\
    \multirow{2}[0]{*}{\model w/o memory} & 39.66 & 68.33 & 67.65 & 76.67 \\
          & (11.98)  & (14.79)  & (8.65)  & (13.05)  \\
          \midrule
    \multirow{2}[1]{*}{\textbf{\model}} & \textbf{54.19} & \textbf{79.17} & \textbf{75.49} & \textbf{85.00} \\
          & \textbf{(11.74)} & \textbf{(14.63)} & \textbf{(8.57)} & \textbf{(12.35)} \\
    \bottomrule
    \end{tabular}}
  \label{tab:ablation}
\end{table}

\begin{figure*}
\vspace{-3mm}
\centering
\subfigure[Case A with 18 original premises.]{
\label{fig:case1}
\includegraphics[width=0.4\textwidth]{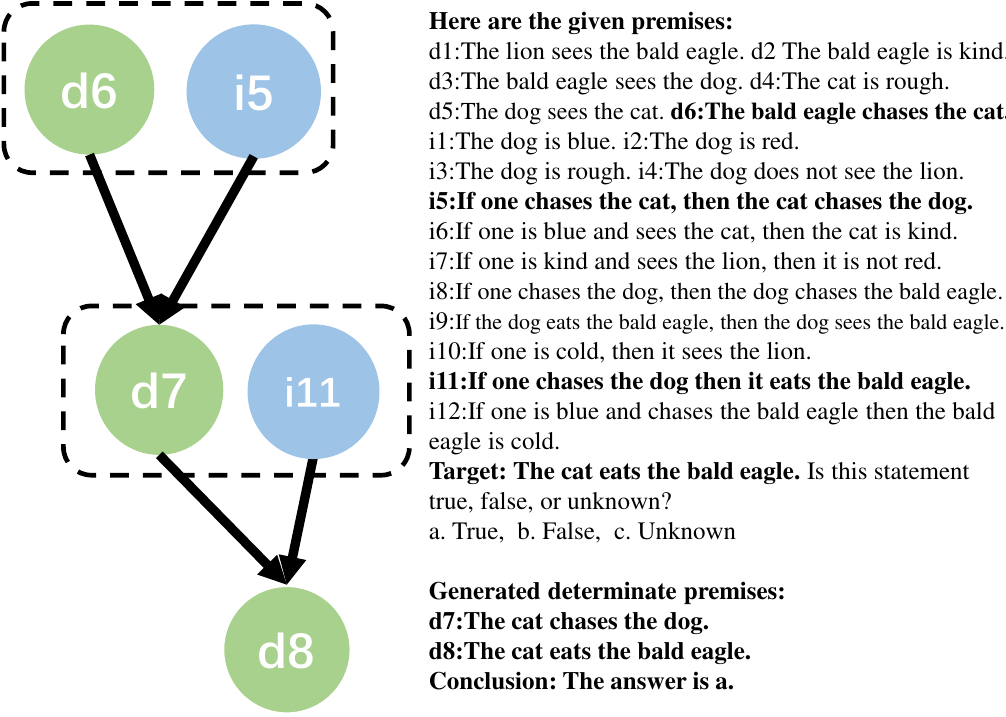}}
\subfigure[Case B with 4 original premises.]{
\label{fig:case2}
\includegraphics[width=0.5\textwidth]{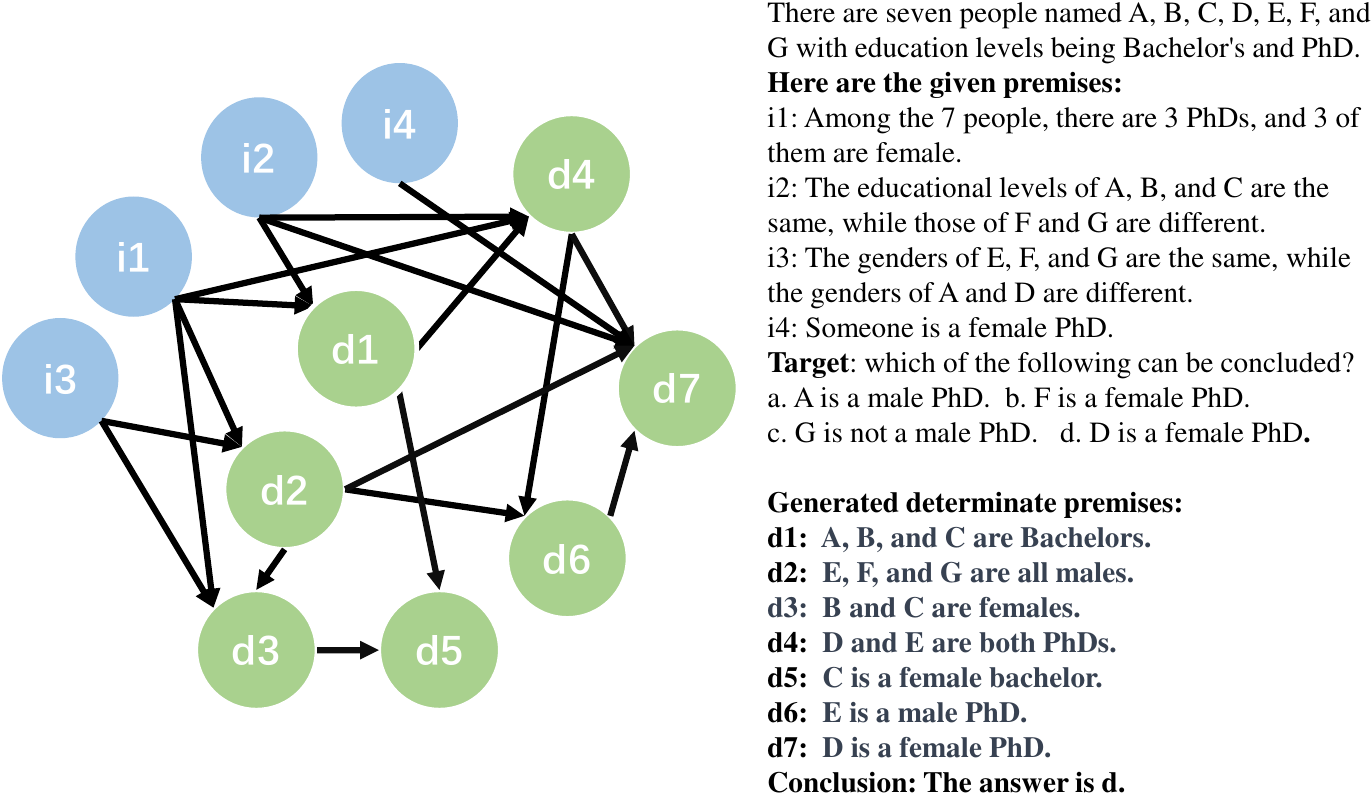}}
\caption{Two cases for contrasting reasoning structure and context complexity. Green dots with $d$ represent determinate premises and blue dots with $i$ represent indeterminate premises.}
\label{fig:case}
\end{figure*}

\subsection{Ablation Study}
In addition to those baselines, we also conduct an ablation study to assess the impact of each component of the proposed method. The ablation variants include: 1) \textbf{\model w/o identify} removes premise identification at the beginning of reasoning; 2) \textbf{\model w/o priority} replaces premise priorities with randomly sampled candidate premises for exploration; 3) \textbf{\model w/o memory} removes our memorization module during iterative reasoning.
The results demonstrate the importance of premise identification, prioritization and exploration, and iterative reasoning memorization modules. 
\model w/o identify blurs the transformation process in reasoning, resulting in reduced accuracy and more reasoning steps.
Since \model w/o priority cannot capture the reasoning direction, it requires more reasoning steps still to achieve a lower accuracy than the full model. This emphasizes that prioritizing premise hierarchy can significantly improve reasoning efficiency.
Without consideration on reasoning memory, the accuracy of \model w/o memory decreases by at least 7.84, indicating the importance of recording the inference structure and prompting LLMs to recall previously acquired information. 

\subsection{Further Analysis}
\paragraph{Case study.}
An intuition suggests that problems with more known conditions and longer contexts tend to require more complex reasoning structures. 
However, relying solely on this intuition to preset reasoning structures might not always be accurate. 
As shown in Figure~\ref{fig:case}, Case A initially appears to be a highly complex problem due to its 18 premises. However, upon prioritizing the premises, we review that the problem's reasoning could be modeled using a concise chain-like reasoning structure with only two steps. 
This indicates the reasoning structure should not be preset before reasoning but rather formed through review after problem solving.
Case B presents only 4 premises, but each premise is complicated and requires to be repeatedly utilized to reach the conclusion.
This also indicates that determining a problem's difficulty solely based on the complexity of its context might not always be accurate.
More reasoning examples and detailed reasoning processes are available in Appendix.

\begin{figure*}
\vspace{-3mm}
\centering
\subfigure[LogiQA.]{
\label{fig:log}
\includegraphics[width=0.4\linewidth]{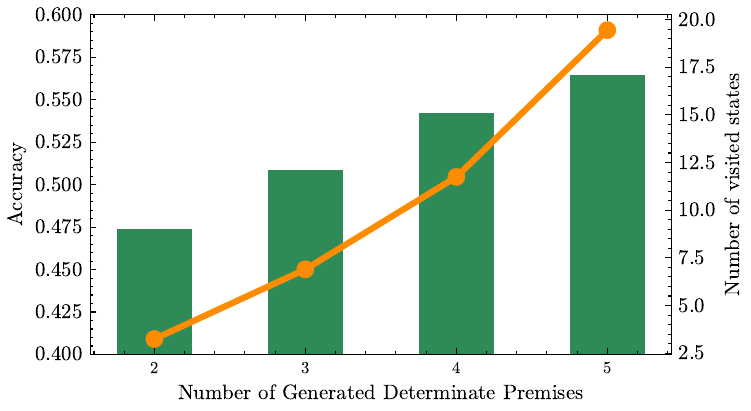}}
\subfigure[FOLIO.]{
\label{fig:fol}
\includegraphics[width=0.4\linewidth]{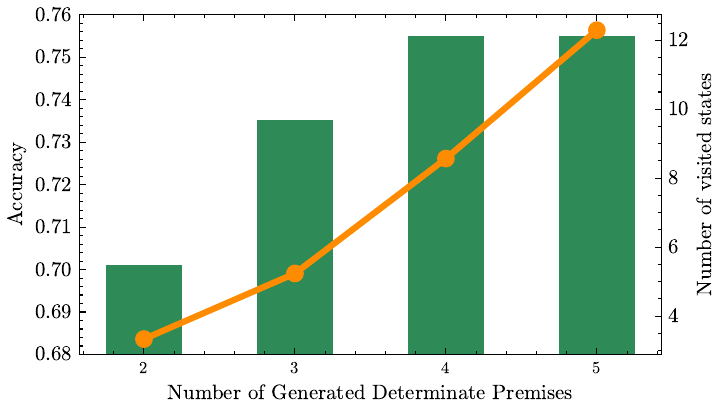}}
\caption{The impact of the number of generated determinate premises.}
\label{fig:number}
\end{figure*}

\paragraph{Impact of the number of determinate premises.}
In practice, the required number of generated determinate premises (denoted as $n$) is a key hyperparameter for our method.
To weigh the reasoning effectiveness and efficiency, a larger number may not always be better.
As shown in Figure~\ref{fig:number}, generating more determinate premises will gradually streamline the reasoning process.
The reasoning performance of smaller $n$ is comparable to some baselines, and as $n$ increases, our method can achieve the best performance. However, the subsequent increase in $n$ will bring about a significant increase in the number of reasoning iterations, while the improvement in accuracy will be very limited. Therefore, we set $n$ to 4 in our experiment, a position close to the inflection point to trade off effect improvement and efficiency control.

\paragraph{Generation efficiency of determinate
premises.}
we further investigate the odds of success and failure across multiple reasoning iterations.
Therefore, we make a more detailed analysis of the average number of reasoning steps required to generate a useful proposition.
As shown in Table~\ref{tab:avgnum}, \model outperforms the baseline significantly across all datasets by using the fewest steps to generate each useful new insight. This number could also reflect the average difficulty of the datasets to some extent. For LogiQA, both CR (4.25) and \model (2.63) are much higher than the values on other datasets, which indicates that the analytical reasoning questions in LogiQA are still the most challenging tasks. 
To validate this hypothesis, we obtain human performance on LogiQA by inviting two volunteers---—one, a graduate student with experience in public examinations, completed all questions with an accuracy rate of 73, and the other, a student with no prior exposure to such knowledge, achieved an accuracy rate of 59.

\paragraph{Complexity analysis.}
We also conduct a complexity analysis for more detailed efficiency comparisons. We choose ToT and CR as strong multi-step reasoning baselines to compute the average inference time for each reasoning step.
All experiments in this analysis are performed on the same device for fair comparison.
The results in Table~\ref{tab:complexity} show that although the inference time per step for \model is slightly more than CR and ToT, the superiority of \model lies in substantially saving overall required reasoning steps. Therefore, considering the inference time per case, we can see that the overall inference efficiency of DetermLR is still better than ToT and CR.


\begin{table}
  \centering
  \caption{The average number of reasoning steps per generated determinate premise.}
  \resizebox{\linewidth}{!}{
    \begin{tabular}{@{}lcccc@{}}
    \toprule
    \textbf{Method} & \textbf{LogiQA} & \textbf{ProofWriter} & \textbf{LD} & \textbf{FOLIO} \\
    \midrule
    ToT & 4.97 & 4.91 & 4.37 & 4.78 \\ 
    CR    & 4.25  & 3.35  & 3.40  & 3.97  \\
    \midrule
    \model w/o identify & 4.31  & 3.32  & 3.37  & 3.43  \\
    \model w/o priority & 4.59  & 3.44  & 3.40  & 3.67  \\
    \model w/o memory & 2.73  & 2.24  & 2.07  & 1.99  \\
    \textbf{\model} & \textbf{2.63} & \textbf{2.17} & \textbf{2.03} & \textbf{1.83} \\
    \bottomrule
    \end{tabular}
    }
  \label{tab:avgnum}
\end{table}


\paragraph{Error analysis.} 
Current LLM-based reasoning cannot resolve the following errors: (1) Insufficient exploration of implicit conditions: LLMs cannot identify that school roommates have the same gender; (2) Insufficient understanding of boundary conditions: Three of the five are candidates, the first two and the last two each have one candidate, LLMs cannot assert that the middle one must be the candidate; (3) Lack of flexible use of logical rules: Given that A implies B,  $\neg$ A implies B, LLMs cannot assert that B must be true.

\begin{table}
  \centering
  \caption{Comparison results of inference efficiency.}
  \resizebox{\linewidth}{!}{
    \begin{tabular}{@{}lccc@{}}
    \toprule
    \multirow{2}[2]{*}{\textbf{Method}} & \textbf{Avg. steps} & \textbf{Inference time} & \textbf{Inference time} \\
          & \textbf{per case} & \textbf{per step} & \textbf{per case} \\
    \midrule
    ToT   & 18.40  & 7.77s & 142.93s \\
    CR    & 14.51  & \textbf{6.86s} & 99.69s \\
    \textbf{DetermLR} & \textbf{7.69 } & 8.05s & \textbf{61.98s} \\
    \bottomrule
    \end{tabular}}
  \label{tab:complexity}
\end{table}

\section{Conclusion}
In this work, we propose \model, a novel reasoning framework to align LLM-based reasoning more closely resemble human cognitive reasoning.
First, we propose a novel perspective that formulates the reasoning process as an evolution from indeterminacy to determinacy. Second, we employ quantitative measurements for premise prioritization and exploration, allowing LLMs to prioritize premises more conducive to exploring new insights.
Furthermore, we introduce a reasoning memorization module to preserve historical details during iterative reasoning. Experimental results show that \model can achieve the highest accuracy on multiple logical reasoning benchmarks while requiring fewer reasoning steps. 
Notably, in more intricate tasks like LogiQA, \model exhibits even more pronounced advancements, mirroring human-like reasoning skills to a greater extent.

\section{Limitation}
While DetermLR exhibits superior performance over baselines across various tasks, challenges persist within LLM-based reasoning. One primary limitation lies in extracting implicit conditions from problem contexts for intricate reasoning tasks. While LLMs can discern intuitively presented conditions, parsing nuanced or implicit information from verbose descriptions remains difficult. In scenarios with few available conditions, exhaustive exploration of each condition is imperative for deriving additional useful conditions. Moreover, LLMs face challenges in accurately parsing exceedingly lengthy condition descriptions, as well as navigating intricate nested if-then conditional structures delineating complex relationships among multiple subjects. Such complexities increase the risk of LLMs erring during the reasoning process. Consequently, enhancing LLMs' proficiency in uncovering implicit conditions and comprehending lengthy and complex condition descriptions emerges as a pivotal research avenue within the domain of LLM-based reasoning. These aspects will be a focal point of our future research directions.

\bibliography{determlr}

\newpage
\appendix

\onecolumn

\section{More Examples on Datasets}
\subsection{LogicalDeduction}
\begin{mdframed}[linewidth=1pt,linecolor=black]
\textbf{[An example in LogicalDeduction]} 

\textbf{Context:} \newline
\hspace{3cm}The following paragraphs each describe a set of seven objects arranged in a fixed order. The statements are logically consistent within each paragraph.In a golf tournament, there were seven golfers: Ana, Eli, Mya, Amy, Joe, Mel, and Ada. Ada finished above Mel. Ada finished third-to-last. Amy finished above Ana. Mya finished second-to-last. Joe finished above Amy. Eli finished below Ana. Ana finished above Ada. Ana finished above Mya.

\textbf{Question:} Which of the following is true?

\textbf{Options:} \newline
A) Ana finished third. \newline
B) Eli finished third. \newline
C) Mya finished third. \newline
D) Amy finished third. \newline
E) Joe finished third. \newline
F) Mel finished third. \newline 
G) Ada finished third.

\textbf{Answer: A} 

\end{mdframed}
\begin{mdframed}[linewidth=1pt,linecolor=black]
\textbf{[Chain-of-Thought Reasoning by GPT-4]}

\textbf{Conclusion:} 
\begin{itemize}
    \item From the premises, we know that: 1. Ada finished third-to-last, which means Ada finished 6th. 2. Mya finished second-to-last, which means Mya finished 6th. 3. Joe finished above Amy. 4. Amy finished above Ana. 5. Eli finished below Ana. 6. Ada finished above Mel. Now, let's try to arrange the golfers based on the given information: 1. Joe 2. Ana 3.  Amy  4. Eli 5. - 6. Mya and Ada 7. Mel From the arrangement, we can see that Amy finished third. So, the answer is D) Amy finished third.
\end{itemize}

\textbf{[Prediction: D]} (\textcolor{red}{Wrong})
\end{mdframed}
\begin{mdframed}[linewidth=1pt,linecolor=black]
\textbf{[Cumulative Reasoning by GPT-4]}

\textbf{Generated Propositions:} 
\begin{itemize}
\item Mel finished last.
\item Ana finished fifth.
\item Eli finished sixth.
\item Mel finished last. Ada finished fifth.
\item Amy finished above Ana and below Joe.
\item Joe finished third.
\end{itemize}

\textbf{Conclusion:} 
\begin{itemize}
    \item From the propositions, we know that Joe finished third. So the answer is E.
\end{itemize}
\textbf{[Prediction: E]} (\textcolor{red}{Wrong})
\end{mdframed}

\newpage
\begin{mdframed}[linewidth=1pt,linecolor=black]
\textbf{[\model by GPT-4]}

\textbf{Generated Determinate Premises:} 
\begin{itemize}
\item Mya did not finish first.
\item Ada did not finish first or second.
\item Joe finished above Ana.
\item Amy did not finish third-to-last.
\item Amy finished above Eli.
\item Amy finished second. Joe finished first.
\item Ada did not finish first-to-third.
\item Eli did not finish first-to-third.
\end{itemize}

\textbf{Reasoning Details:} 
\begin{enumerate}
\item In the NO:1 round, we use these ``premises'': ``Amy finished above Ana. Mya finished second-to-last. Ada finished above Mel. Eli finished below Ana. Joe finished above Amy.'' and got a ``New Determinate Premise'': ``Joe finished above Ana''
\item In the NO:2 round, we use these ``premises'': `` Amy finished above Ana. Eli finished below Ana.'' and got a ``New Determinate Premise'': ``Amy finished above Eli.''
\item In the NO:3 round, we use these ``premises'': `` Ada finished third-to-last. Amy finished above Ana.'' and got a ``New Determinate Premise'': ``Amy did not finish third-to-last.''
\item In the NO:4 round, we use these ``premises'': ``Amy finished first or second. Joe finished above Amy. '' and got a ``New Determinate Premise'': ``Amy finished second. Joe finished first.''
\item In the NO:5 round, we use these ``premises'': `` Ada finished third-to-last. Ada finished above Mel. '' and got a ``New Determinate Premise'': ``Ada did not finish first-to-third. ''
\item In the NO:6 round, we use these ``premises'': `` Eli finished below Ana. Ana did not finish first or second. '' and got a ``New Determinate Premise'': ``Eli did not finish first-to-third. ''
\end{enumerate}

\textbf{Conclusion:} 
\begin{itemize}
    \item From the premises and propositions, we know that Ada finished third-to-last, Ana finished above Ada, Ana finished above Mya, which means she finished fifth. Mya finished second-to-last, which means she finished sixth. Since there are only seven golfers, this means that Joe must have finished first, as he finished above Amy, who finished above Ana, who finished above Eli. With Joe finishing first, Amy must have finished second, since she finished above Ana and Eli. Ana would be third, Eli would be fourth, and Mel would be the last (seventh). So the answer is A) Ana finished third.
\end{itemize}
\textbf{[Prediction: A]} (\textcolor{green}{Right})
\end{mdframed}
\newpage
\subsection{LogiQA}
\begin{mdframed}[linewidth=1pt,linecolor=black]
\textbf{[An example in LogiQA]}

\textbf{Context:} \newline
\hspace{3cm}There are seven people named A, B, C, D, E, F, and G with education levels being Bachelor's and PhD.
Here are the given premises: 1): Among the 7 individuals, there are 3 with a PhD., and 3 of them are female.
2) The educational levels of A, B, and C are the same, while those of F and G are different.
3) The genders of E, F, and G are the same, while the genders of A and D are different.
4) The final hire is a female PhD.

\textbf{Question: }Based on the statements above, which of the following can be concluded?

\textbf{Options:} \newline
A. A is a male PhD.\newline
B. F is a female PhD.\newline
C. G is not a male PhD. \newline
D. D is a female PhD.

\textbf{Answer: D}
\end{mdframed}

\begin{mdframed}[linewidth=1pt,linecolor=black]
\textbf{[Chain-of-Thought Reasoning by GPT-4]} 

\textbf{Conclusion:} 
\begin{itemize}
    \item From the premise, it can be inferred that the genders of E, F, and G are female. Since the final appointment is a female Ph.D. holder, the selection must be made from E, F, and G. The educational levels of A, B, and D are the same, so they cannot be Ph.D. holders because there are only 3 Ph.D. holders. Therefore, there must be two Ph.D. holders among E, F, and G. Since the educational levels of F and G are different, one of them must be a Ph.D. holder, and E is also a Ph.D. holder. Therefore, one of E, F, and G is a female Ph.D. holder.
The genders of A and C are different, and since E, F, and G are females, either A or C must be male. Among options A, B, C, and D, option B is correct because F is a female Ph.D. holder. The answer is B.
\end{itemize}

\textbf{[Prediction: B]} (\textcolor{red}{Wrong})
\end{mdframed}
\begin{mdframed}[linewidth=1pt,linecolor=black]
\textbf{[Cumulative Reasoning by GPT-4]} \newline
\textbf{Generated Propositions} 
\begin{itemize}
\item The genders of D,E,F are female.
\item A,B,C are undergraduates.
\end{itemize}

\textbf{Conclusion:} 
\begin{itemize}
    \item From the proposition, we can deduce that E, F, and G are females. A and C have different genders, so A and C are males. A, B, and D have the same educational background, which means they are either all Ph.D. holders or all undergraduates. However, there are only 3 Ph.D. holders, so A, B, and D must all be undergraduates. This means that E, F, and G include two Ph.D. holders. Since the final appointment is a female Ph.D. holder, it means that either E or F must be a Ph.D. holder. However, the educational levels of F and G are different, so G is an undergraduate. Therefore, both E and F are Ph.D. holders. Therefore, the final answer is: A is a male undergraduate; C is a male undergraduate; E is a female Ph.D. holder; F is a female Ph.D. holder; G is a female undergraduate. The answer is C.
\end{itemize}

\textbf{[Prediction: C]} (\textcolor{red}{Wrong})
\end{mdframed}

\newpage
\begin{mdframed}[linewidth=1pt,linecolor=black]
\textbf{[\model by GPT-4]}

\textbf{Generated Determinate Premises:} 
\begin{itemize}
\item A, B, and C have Bachelor's degrees.
\item E, F, and G are all males.
\item B and C are females.
\item D and E are both PhD. holders.
\item C is a female undergraduate.
\item E is a male PhD. holder.
\item D is a female PhD. holder.
\end{itemize}

\textbf{Reasoning Details:}
\begin{enumerate}
\item In the NO:1 round, we use these ``premises'': ``Among the 7 individuals, there are 3 with a PhD., and 3 of them are female. The educational levels of A, B, and C are the same, while those of F and G are different.'' and got a ``New Determinate Premise'': ``A, B, and C have Bachelor's degrees.''
\item In the NO:2 round, we use these ``premises'': `` Among the 7 individuals, there are 3 with a PhD., and 3 of them are female. The genders of E, F, and G are the same, while the genders of A and D are different.'' and got a ``New Determinate Premise'': ``E, F, and G are all males.''
\item In the NO:3 round, we use these ``premises'': `` Among the 7 individuals, there are 3 with a PhD., and 3 of them are female. The genders of E, F, and G are the same, while the genders of A and D are different. E, F, and G are all males'' and got a ``New Determinate Premise'': ``B and C are females.''
\item In the NO:4 round, we use these ``premises'': ``Among the 7 individuals, there are 3 with a PhD., and 3 of them are female. The genders of E, F, and G are the same, while the genders of A and D are different. A, B, and C have Bachelor's degrees.'' and got a ``New Determinate Premise'': ``D and E are both PhD. holders.''
\item In the NO:5 round, we use these ``premises'': ``B and C are females. A, B, and C have Bachelor's degrees.'' and got a ``New Determinate Premise'': ``C is a female undergraduate.''
\item In the NO:6 round, we use these ``premises'':`` D and E are both PhD. Holders. E, F, and G are all males '' and got a ``New Determinate Premise'': ``E is a male PhD. holder.''
In the NO:7 round, we use these ``premises'': ``The educational levels of A, B, and C are the same, while those of F and G are different. The final hire is a female PhD. E is a male PhD. holder. ''and got a ``New Determinate Premise'': `` D is a female PhD. holder''
\end{enumerate}

\textbf{Conclusion:} 
\begin{itemize}
    \item From the propositions, we know the answer is D. D is a female PhD.
\end{itemize}
\textbf{[Prediction: D]} (\textcolor{green}{Right})
\end{mdframed}

\newpage
\subsection{ProofWriter}
\begin{mdframed}[linewidth=1pt,linecolor=black]
\textbf{[An example in ProofWriter]} 

\textbf{Context:} \newline
\hspace{3cm}The bald eagle chases the cat. The bald eagle is kind. The bald eagle sees the dog. The cat is rough. The dog is blue. The dog is red. The dog is rough. The dog sees the cat. The dog does not see the lion. The lion sees the bald eagle. If something chases the cat then the cat chases the dog. If something chases the dog then it eats the bald eagle. If something is kind and it sees the lion then it is not red. If something chases the dog then the dog chases the bald eagle. If the dog eats the bald eagle then the dog sees the bald eagle. If something is cold then it sees the lion. If something sees the cat and it is blue then the cat is kind. If something is blue and it chases the bald eagle then the bald eagle is cold.

\textbf{Question:}Based on the above information, is the following statement true, false, or uncertain? The cat eats the bald eagle.

\textbf{Options:} \newline
A) True\newline
B) False   \newline
C) Uncertain

\textbf{Answer: A}
\end{mdframed}

\begin{mdframed}[linewidth=1pt,linecolor=black]
\textbf{[Chain-of-Thought Reasoning by GPT-4]}

\textbf{Conclusion:} 
\begin{itemize}
    \item The bald eagle chases the cat.2. If something chases the cat, then the cat chases the dog. (From 1, the cat chases the dog)3. If something chases the dog, then it eats the bald eagle. (We don‘t have information about something chasing the dog) 4. The cat eats the bald eagle. (We don’t have information about the cat eating the bald eagle)Recall the Hypothesis: The cat eats the bald eagle is Unknown.
\end{itemize}

\textbf{[Prediction: C]} (\textcolor{red}{Wrong})
\end{mdframed}
\begin{mdframed}[linewidth=1pt,linecolor=black]
\textbf{[Cumulative Reasoning by GPT-4]} 

\textbf{Generated Propositions:} 
\begin{itemize}
\item The bald eagle does not eat the dog.
\item If the bald eagle sees the lion, it is not red.
\item The bald eagle is not red.
\item If the dog chases the cat, then the cat chases the dog.
\item The dog does not eat the bald eagle.
\end{itemize}

\textbf{Conclusion:} 
\begin{itemize}
    \item We can not find any direct or indirect information about the cat eating the bald eagle. The information provided in the premises does not support or contradict the hypothesis. Judgement: Now we know that the Hypothesis is Unknown.
\end{itemize}

\textbf{[Prediction: C]} (\textcolor{red}{Wrong})
\end{mdframed}

\newpage
\begin{mdframed}[linewidth=1pt,linecolor=black]
\textbf{[\model Reasoning by GPT-4]}\newline
\textbf{Generated Determinate Premises:} 
\begin{enumerate}
\item The cat chases the dog.
\item The dog chases the bald eagle.
\item Something chases the cat.
\item The bald eagle is something that chases the cat.
\item The dog and the lion both see the bald eagle.
\item If the cat chases the dog, then the cat eats the bald eagle.
\item The cat eats the bald eagle.
\end{enumerate}
\textbf{Reasoning Details: }
\begin{itemize}
\item In the NO:1 round, we use these ``premises'': ``If something chases the cat then the cat chases the dog. The bald eagle chases the cat.''and got a ``New Determinate Premise'': ``The cat chases the dog.''
\item In the NO:2 round, we use these ``premise'': ``If something chases the dog then the dog chases the bald eagle. If something chases the cat then the cat chases the dog. The dog sees the cat. The cat chases the dog. If something chases the dog then it eats the bald eagle. The bald eagle chases the cat. The cat is rough.''and got a ``New Determinate Premise'': ``The dog chases the bald eagle.''
 \item In the NO:3 round, we use these ``premises'': ``If something chases the cat then the cat chases the dog. The cat chases the dog. If something chases the dog then it eats the bald eagle. The bald eagle chases the cat.''and got a ``New Determinate Premise'': ``Something chases the cat.''
 \item In the NO:4 round, we use these ``premises'': ``Something chases the cat. The bald eagle chases the cat.''and got a ``New Determinate Premise'': ``The bald eagle is something that chases the cat.''
 \item In the NO:5 round, we use these ``premises'': ``The dog chases the bald eagle. The bald eagle is something that chases the cat. The bald eagle chases the cat. The bald eagle is kind. The bald eagle sees the dog. The lion sees the bald eagle.''and got a ``New Determinate Premise”: “The dog and the lion both see the bald eagle.''
 \item In the NO:6 round, we use these ``premises'': ``The dog chases the bald eagle. The bald eagle sees the dog. If something chases the dog then the dog chases the bald eagle. If something chases the dog then it eats the bald eagle. The dog is blue.'' and got a ``New Determinate Premise'': ``If the cat chases the dog, then the cat eats the bald eagle.''
 \item In the NO:7 round, we use these ``premises'': ``The cat chases the dog. The bald eagle is something that chases the cat. If something chases the dog then it eats the bald eagle. If something chases the cat then the cat chases the dog. If the cat chases the dog, then the cat eats the bald eagle. The dog chases the bald eagle. The bald eagle chases the cat.''and got a ``New Determinate Premise'': ``The cat eats the bald eagle.''
 \end{itemize}
\textbf{Conclusion:} 
    From the propositions, we know the cat eats the bald eagle. So the answer is true. \newline
\textbf{[Prediction: A]} (\textcolor{green}{Right})
\end{mdframed}

\newpage
\subsection{FOLIO}
\begin{mdframed}[linewidth=1pt,linecolor=black]
\textbf{[An example in FOLIO]}

\textbf{Context:} \newline
\hspace{3cm}No digital media are analog. Every printed text is analog. All streaming services are digital. If an object is a hardcover book then it is printed text. If 1984 is a streaming service, then 1984 is a hardcover book.

\textbf{Question:} Based on the above information, is the following statement true, false, or uncertain? 1984 is not a streaming service.

\textbf{Options:} \newline
A) True\newline
B) False   \newline
C) Uncertain\newline
\textbf{Answer: A}
\end{mdframed}

\begin{mdframed}[linewidth=1pt,linecolor=black]
\textbf{[Chain-of-Thought Reasoning by GPT-4]}

\textbf{Conclusion:} 
\begin{itemize}
    \item From the premises, we cannot directly deduce whether the hypothesis 1984 is not a streaming service is true or false. The premises provide information about the different media formats, but there is no explicit information about the format of 1984. So the answer is C.
\end{itemize}

\textbf{[Prediction: C]} (\textcolor{red}{Wrong})
\end{mdframed}
\begin{mdframed}[linewidth=1pt,linecolor=black]
\textbf{[Cumulative Reasoning by GPT-4]} 

\textbf{Generated Propositions:} 
\begin{itemize}
\item If 1984 is a streaming service, then it is digital.
\item If 1984 is a streaming service, then it is not analog.
\end{itemize}

\textbf{Conclusion:} 
\begin{itemize}
    \item The premises state that if 1984 is a streaming service, then it is a hardcover book. However, no information is provided about whether 1984 is a streaming service or a hardcover book. Since we lack information to confirm or deny the hypothesis, we cannot deduce a conclusion from the given premises. Judgement: Now we know that the Hypothesis is Unknown. So the answer is C.
\end{itemize}
\textbf{[Prediction: C]} (\textcolor{red}{Wrong})
\end{mdframed}

\newpage
\begin{mdframed}[linewidth=1pt,linecolor=black]
\textbf{[\model Reasoning by GPT-4]} 

\textbf{Generated Determinate Premises:} 
\begin{itemize}
\item If 1984 is a streaming service, then 1984 is digital.
\item All streaming services are not analog.
\item If 1984 is a streaming service, then it cannot be a hardcover book.
\end{itemize}
\textbf{Reasoning Details: }
\begin{enumerate}
\item In the NO:1 round, we use these ``premises'': ``If 1984 is a streaming service, then 1984 is a hardcover book. All streaming services are digital.'' and got a ``New Determinate Premise'': ``If 1984 is a streaming service, then 1984 is digital.''
\item In the NO:2 round, we use these ``premises'': ``No digital media are analog. All streaming services are digital. If 1984 is a streaming service, then 1984 is a hardcover book. If 1984 is a streaming service, then 1984 is digital.'' and got a ``New Determinate Premise'': ``All streaming services are not analog.''
\item In the NO:3 round, we use these ``premises'': ``If 1984 is a streaming service, then 1984 is digital. If 1984 is a streaming service, then 1984 is a hardcover book. All streaming services are digital. No digital media are analog. All streaming services are not analog.'' and got a ``New Determinate Premise'': ``If 1984 is a streaming service, then it cannot be a hardcover book.''
\end{enumerate}

\textbf{Conclusion:} 
\begin{itemize}
    \item We know from the premises that if 1984 is a streaming service, then 1984 is a hardcover book. However, we also deduced that if 1984 is a streaming service, then it cannot be a hardcover book, as it would be digital and not analog. This contradiction leads us to conclude that 1984 is not a streaming service. Judgement: Now we know that the Hypothesis is True.  
\end{itemize}
\textbf{[Prediction: A]} (\textcolor{green}{Right})
\end{mdframed}

\newpage
\section{Prompt Templates}
Different logical reasoning tasks have different data formats. Among them, ProofWriter and FOLIO offer explicitly known premises, while LogiQA and LogicalDeduction require parsing the question context to extract the premises, which indicates that extra steps such as premise transformation are necessary.
We use ProofWriter and LogicalDeduction as representative tasks to illustrate the prompt templates. More details about prompt design are available in our Github repository.

\subsection{ProofWriter}
Based on the modeling scheme introduced by our \model, we summarize main designed prompts into several parts such as premise identification, premise prioritization, premise exploration, logical validation, and final conclusion.
\begin{mdframed}[linewidth=1pt,linecolor=black]
\textbf{Prompts used for Premise Identification}\\
\{\{\#system\}\}Suppose you are one of the greatest AI scientists, logicians and mathematicians. \newline
Let us think step by step. \newline
First, read and analyze the following definition:\newline
Determinate premise: The premise contains the same noun or adjective as the Hypothesis, and the premise is not in the structure of ``if...'' or ``if...then...''.\newline
Second, read and analyze the ``Premise'' and ``Hypothesis'' .Judge ``Premise'' is ``determinate premise'' or not.\newline
Third, please make sure your classification decisions are derived directly from definitions, rather than unsourced common sense.\newline
---\newline
\{\{\/system\}\}\newline
\{\{\~\#each examples\}\}\newline
\{\{\#user\}\}\newline
---\newline
``Premise'': ``\{\{this.Premise\}\}''\newline
``Hypothesis'': ``\{\{this.Hypothesis\}\}''\newline
\{\{/user\}\}\newline
\raggedright\{\{\#assistant\}\}``Judgement'':``Is this ''Premise`` a ''determinate premise`` or not?
\{\{this.usefulness\}\}'' \{\{/assistant\}\}\newline
\{\{\#assistant\}\}``Explanation'': \{\{this.Explanation\}\}\{\{\/assistant\}\}\newline
\{\{\~/each\}\}\newline
\end{mdframed}

\newpage
\begin{mdframed}[linewidth=1pt,linecolor=black]
\textbf{Prompts used for Premise Prioritization}\\
\{\{\#system\}\}Suppose you are one of the greatest AI scientists, logicians and mathematicians. Let us think step by step. 
Read and analyze the ``determinate premise'' and ``indeterminate premise''\\
first, then selecting several premises from them. \\
Read the ``Last reasoning history''.If we got a ``false Proposition'' in history,when you select ``Most relevant premise'',do not choose the same ``Most relevant premise'' in history as your answer.\\
Please follow these steps:\\
1.From the determinate premise, select the ``Most relevant premise'' which has the same subject with ``Hypothesis'', and give a score from 0 to 1.\\
2.You need to assess how the ``Most relevant premise'' relates to all the other ``determinate premise'' and ``indeterminate premise'',based on Relevance scoring rules.\\
3.The ``determinate premise'' and ``indeterminate premise'' with scores higher than 0.25 will be used as the final results, along with Most relevant premise.\\
Relevance scoring rules:\\
1. When scoring relevance, 0.25 added for each noun or 0.3 added for each adjective that is the same between two sentences.\\
2. Scores start to accumulate from 0 points, and the upper limit is 1 point.\\
3. If sentence p1 is a hypothetical premise of sentence p2,then add 0.25 to p2. for example: measure ``if A then B.'' and ``A is true.'' Then add 0.25 points to ``if A then B''.\\
----\\
\{\{\/system\}\}\\
\{\{\~\#each examples\}\}\\
\{\{\#user\}\}\\
---\\
``determinate premise'': ``\{\{this.determinate premise\}\}''\\
``indeterminate premise'': ``\{\{this.indeterminate premise\}\}''\\
``Hypothesis'': ``\{\{this.Hypothesis\}\}''\\
``Last reasoning history'': ``\{\{this.last history\}\}''\\
\{\{\/user\}\}\\
\{\{\#assistant\}\}Can you select the premise from the ``determinate premises'' that scores the highest score for Relevance scoring rules to the ``hypothesis''?\{\{\/assistant\}\}\\
\{\{\#assistant\}\}``Most relevant premise'': ``\{\{this.Most relevant premise\}\}''\{\{\/assistant\}\}\\
\{\{\#assistant\}\}Can you assess how the ``Most relevant premise'' relates to all the other ``determinate premise'' and ``indeterminate premise'' accoding to Relevance scoring rules?\{\{\/assistant\}\}\\
\{\{\#assistant\}\}``Other premises scores'': ``\{\{this.Other premises scores\}\}''\{\{\/assistant\}\}\\
\{\{\#assistant\}\}``Results'': ``\{\{this.Results\}\}''\{\{\/assistant\}\}\\
\{\{\~\/each\}\}
\end{mdframed}

\newpage
\begin{mdframed}[linewidth=1pt,linecolor=black]
\textbf{Prompts used for Premise Exploration}\\
\{\{\#system\}\}Suppose you are one of the greatest AI scientists, logicians and mathematicians. Let us think step by step. \\
Please use Logical Reasoning Rules(LRR) to deduce a ``Proposition'' from two given ``Premises'' and the proposition does not include ``if''.\\
Logical Reasoning Rules(LRR):\\
1. ``Two premises'': ``If A,then B. A is true.'' then ``Proposition'': ``B is true.''\\
2. ``Two premises'': ``If A,then B. B is not true.'' then ``Proposition'': ``A is not true''\\
3. ``Two premises'': ``A is either C or D. A is not C.'' then ``Proposition'': ``A is D.''\\
Please make sure that the ``Proposition'' is logically correct.\\
Please make sure that the ``Proposition'' is not a duplicate of the ``Premises''.\\
Please make sure your reasoning is directly deduced from the ``Premises'' and ``Propositions'' other than introducing unsourced common knowledge and unsourced information by common sense reasoning.\\
Please remember that your ``Proposition'' should be useful to determine whether the ``Hypothesis'' is True, False or Unknown.\\
----\\
\{\{\/system\}\}\\

\{\{\~\#each examples\}\}\\
\{\{\#user\}\}\\
---\\
``Premises'': ``\{\{this.premises\}\}''\\
We want to deduce more propositions to determine the correctness of the following ``Hypothesis'':\\
``Hypothesis'': ``\{\{this.conclusion\}\}''\\
Can you deduce a new ``Proposition'' from at least two given ``Premises''?\\
\{\{\/user\}\}\\

\{\{\#assistant\}\}``Proposition'': ``\{\{this.proposition\}\}''\{\{\/assistant\}\}\\
\{\{\~\/each\}\}\\
\{\{\#user\}\}\\
---\\
``premises'': ``\{\{premises\}\}''\\
``boundary condition'': ``\{\{boundary condition\}\}''\\
We want to derive more propositions to solve the following question:\\
``question'': ``\{\{question\}\}''\\
Combined with boundary conditions, can you derive a new ``proposition'' from at least two given ``premises''?\\
\{\{\/user\}\}\\

\{\{\#assistant\}\}``proposition'': ``\{\{\/assistant\}\}\\
\{\{\#assistant\}\}\{\{gen ''proposition`` temperature=temperature max tokens=100 stop=' '\}\}\{\{\/assistant\}\}\\
\end{mdframed}

\newpage
\begin{mdframed}[linewidth=1pt,linecolor=black]
\textbf{Prompts used for Logical Validity}\\
\{\{\#system\}\}Suppose you are one of the greatest AI scientists, logicians and mathematicians. Let us think step by step. \\Please use the Logical Reasoning Rules(LRR) to determine whether the deduction of the given ``Premises'' to a ``Proposition'' is valid or not, reply with True or False.\\Logical Reasoning Rules(LRR):\\1. ``Two premises'': ``If A,then B. A is true.'' then ``Proposition'': ``B is true.''\\2. ``Two premises'': ``If A,then B. If B,then C.'' then ``Proposition'': ``If A, then C.''\\3. ``Two premises'': ``If A,then B. B is not true.'' then ``Proposition'': ``A is not true''\\4. ``Two premises'': ``A is either C or D. A is not C.'' then ``Proposition'': ``A is D.''\\---\\\{\{\/system\}\}\\\{\{\~\#each examples\}\}\\\{\{\#user\}\}\\---\\``Premises'': ``\{\{this.premises\}\}''\\``Proposition'': ``\{\{this.proposition\}\}''\\\{\{\/user\}\}\\\{\{\#assistant\}\}``Judgement'': ``Is this deduction valid? \{\{this.validation\}\}''\{\{\/assistant\}\}\\\{\{\~\/each\}\}
\end{mdframed}

\newpage
\begin{mdframed}[linewidth=1pt,linecolor=black]
\textbf{Prompts used for Final Conclusion}\\
\{\{\#system\}\}Suppose you are one of the greatest AI scientists, logicians, and mathematicians. Let's think about it step by step.\\
First read and analyze the ``paragraphs'' and ``questions'', then use the ``premises'', ``boundary conditions'' and ``propositions'' to reason which of the options given is the answer to the ``question''.\\
Make sure that your reasoning is derived directly from ``premises'' and ``propositions'' rather than introducing passive common sense and passive information through common sense reasoning.\\
Please note that this is a single choice question.\\
If you can get the answer directly from the proposition, then you should choose the answer directly, otherwise keep reasoning with the proposition, premises, and boundary conditions until you arrive at a single answer.\\
---\\
\{\{\/system\}\}\\
\{\{\~\#each examples\}\}\\
\{\{\#user\}\}\\
---\\
``context'': ``\{\{context\}\}''\\
``question and options'': ``\{\{question\}\}''\\
\{\{\/user\}\}\\
\{\{\#assistant\}\}``Premises'': ``Let's think step by step, and from the context we can extract these premises: \{\{premises\}\}''\{\{\/assistant\}\}\\
\{\{\#assistant\}\}``Boundary condition'': ``Let's think step by step, and from the context we can extract these boundary conditions: \{\{boundary condition\}\}''\{\{\/assistant\}\}\\
\{\{\#assistant\}\}``Thoughts'': ``Let us think step by step. From the premises, we can deduce propositions:\{\{propositions\}\}''\{\{\/assistant\}\}\\
\{\{\#assistant\}\}``Recall the reasoning history'':``\{\{infer history\}\}''\{\{\/assistant\}\}\\
\{\{\#assistant\}\}``Recall the questions and options'':``\{\{question\}\}''\{\{\/assistant\}\}\\
\{\{\#assistant\}\}``Reasoning'': ``Using premises, boundary conditions, and continuing to reason according to the propositions already obtained,\{\{\/assistant\}\}\\
\{\{\#assistant\}\}\{\{gen ''reasoning`` temperature=0.7 max tokens=500 stop=[' ']\}\}\{\{\/assistant\}\}\\
\{\{\#assistant\}\}''Recall the questions and options``:''\{\{question\}\}``\{\{\/assistant\}\}\\
\{\{\#assistant\}\}''Judgement``: ''Now we know that the answer to this question should be\{\{\/assistant\}\}\\
\{\{\#assistant\}\}\{\{select ``judgement'' options=choose\}\}\{\{\/assistant\}\}\\
\end{mdframed}

\newpage
\subsection{LogicalDeduction}
In addition to the prompting steps mentioned above, we also include premise extraction and premise transformation to parse the available premises from the original question.
\begin{mdframed}[linewidth=1pt,linecolor=black]
\textbf{Prompts used for Premise Identification}\\
\{\{\#system\}\}Suppose you are one of the greatest AI scientists, logicians and mathematicians. \newline
Let us think step by step. \newline
First, read and analyze the following definition:\newline
Determinate premise: The premise contains the same noun or adjective as the Hypothesis, and the premise is not in the structure of ``if...'' or ``if...then...''.\newline
Second, read and analyze the ``Premise'' and ``Hypothesis'' .Judge ``Premise'' is ``determinate premise'' or not.\newline
Third, please make sure your classification decisions are derived directly from definitions, rather than unsourced common sense.\newline
---\newline
\{\{\/system\}\}\newline
\{\{\~\#each examples\}\}\newline
\{\{\#user\}\}\newline
---\newline
``Premise'': ``\{\{this.Premise\}\}''\newline
``Hypothesis'': ``\{\{this.Hypothesis\}\}''\newline
\{\{/user\}\}\newline
\raggedright\{\{\#assistant\}\}``Judgement'':``Is this ''Premise`` a ''determinate premise`` or not?
\{\{this.usefulness\}\}'' \{\{/assistant\}\}\newline
\{\{\#assistant\}\}``Explanation'': \{\{this.Explanation\}\}\{\{\/assistant\}\}\newline
\{\{\~/each\}\}\newline
\end{mdframed}

\newpage
\begin{mdframed}[linewidth=1pt,linecolor=black]
\textbf{Prompts used for Premise Prioritization}\\
\{\{\#system\}\}Suppose you are one of the greatest artificial intelligence scientists, logicians, and mathematicians. Let's think about it step by step.\\
First read and analyze the ``determinate premises'' and ``indeterminate premises'', and then filter out several premises.\\
When you decide on a variable, read through the inference history first and don't choose a variable that has failed before as your choice for this round.\\
Please follow these steps:\\
1. Count the cumulative number of times each variable is mentioned by ``determinate premises'' and ``indeterminate premises''.\\
2. Determine the variable according to the number of mentions from high to low. If the number of mentions is the same, the variable with more prerequisites will be given priority.\\
3. Determine whether the value of the variable has been determined under the current variable. If it is determined, search and determine the next variable in order from most to least. If it has not been completely determined, go to step 4.\\
4. Use this variable as a criterion for screening ``premises'' and filter out all premises related to this variable.\\
---\\
\{\{\/system\}\}\\
\{\{\~\#each examples\}\}\\
\{\{\#user\}\}\\
---\\
``determinate premise'': ``\{\{determinate premise\}\}''\\
``indeterminate premise'': ``\{\{indeterminate premise\}\}''\\
``topic'': ``\{\{topic\}\}''\\
``boundary condition'': ``\{\{boundary condition\}\}''\\
``Inference history'': ``\{\{last false history\}\}''\\
\{\{\/user\}\}\\
\{\{\#assistant\}\}Can you count the cumulative number of times each variable is mentioned by the premises?\{\{\/assistant\}\}\\
\{\{\#assistant\}\}``Count'': ``\{\{\/assistant\}\}\\
\{\{\#assistant\}\}\{\{gen ''count`` temperature=temperature max tokens=200 stop=' '\}\}\{\{\/assistant\}\}\\
\{\{\#assistant\}\}Which variable should you choose as the criterion for premises screening?\{\{\/assistant\}\}\\
\{\{\#assistant\}\}''Explanation``: ''\{\{\/assistant\}\}\\
\{\{\#assistant\}\}\{\{gen ``explanation'' temperature=temperature max tokens=200 stop=' '\}\}\{\{\/assistant\}\}\\
\{\{\#assistant\}\}What are all the premises related to this variable?\{\{\/assistant\}\}
\{\{\#assistant\}\}``Results'': ``\{\{\/assistant\}\}\\
\{\{\#assistant\}\}\{\{gen ''results`` temperature=temperature max tokens=200 stop=' '\}\}\{\{\/assistant\}\}
\end{mdframed}

\newpage
\begin{mdframed}[linewidth=1pt,linecolor=black]
\textbf{Prompts used for Premise Exploration}\\
\{\{\#system\}\}Suppose you are one of the greatest AI scientists, logicians and mathematicians. Let us think step by step. \\
Please use Logical Reasoning Rules(LRR) to deduce a ``Proposition'' from two given ``Premises'' and the proposition does not include ``if''.\\
Logical Reasoning Rules(LRR):\\
1. ``Two premises'': ``If A,then B. A is true.'' then ``Proposition'': ``B is true.''\\
2. ``Two premises'': ``If A,then B. B is not true.'' then ``Proposition'': ``A is not true''\\
3. ``Two premises'': ``A is either C or D. A is not C.'' then ``Proposition'': ``A is D.''\\
Please make sure that the ``Proposition'' is logically correct.\\
Please make sure that the ``Proposition'' is not a duplicate of the ``Premises''.\\
Please make sure your reasoning is directly deduced from the ``Premises'' and ``Propositions'' other than introducing unsourced common knowledge and unsourced information by common sense reasoning.\\
Please remember that your ``Proposition'' should be useful to determine whether the ``Hypothesis'' is True, False or Unknown.\\
----\\
\{\{\/system\}\}\\

\{\{\~\#each examples\}\}\\
\{\{\#user\}\}\\
---\\
``Premises'': ``\{\{this.premises\}\}''\\
We want to deduce more propositions to determine the correctness of the following ``Hypothesis'':\\
``Hypothesis'': ``\{\{this.conclusion\}\}''\\
Can you deduce a new ``Proposition'' from at least two given ``Premises''?\\
\{\{\/user\}\}\\

\{\{\#assistant\}\}``Proposition'': ``\{\{this.proposition\}\}''\{\{\/assistant\}\}\\
\{\{\~\/each\}\}\\
\{\{\#user\}\}\\
---\\
``premises'': ``\{\{premises\}\}''\\
``boundary condition'': ``\{\{boundary condition\}\}''\\
We want to derive more propositions to solve the following question:\\
``question'': ``\{\{question\}\}''\\
Combined with boundary conditions, can you derive a new ``proposition'' from at least two given ``premises''?\\
\{\{\/user\}\}\\

\{\{\#assistant\}\}``proposition'': ``\{\{\/assistant\}\}\\
\{\{\#assistant\}\}\{\{gen ''proposition`` temperature=temperature max tokens=100 stop=' '\}\}\{\{\/assistant\}\}\\
\end{mdframed}

\newpage
\begin{mdframed}[linewidth=1pt,linecolor=black]
\textbf{Prompts used for Logical Validation}\\
\{\{\#system\}\}Suppose you are one of the greatest AI scientists, logicians and mathematicians. Let us think step by step. \\Please use the Logical Reasoning Rules(LRR) to determine whether the deduction of the given ``Premises'' to a ``Proposition'' is valid or not, reply with True or False.\\Logical Reasoning Rules(LRR):\\1. ``Two premises'': ``If A,then B. A is true.'' then ``Proposition'': ``B is true.''\\2. ``Two premises'': ``If A,then B. If B,then C.'' then ``Proposition'': ``If A, then C.''\\3. ``Two premises'': ``If A,then B. B is not true.'' then ``Proposition'': ``A is not true''\\4. ``Two premises'': ``A is either C or D. A is not C.'' then ``Proposition'': ``A is D.''\\---\\\{\{\/system\}\}\\\{\{\~\#each examples\}\}\\\{\{\#user\}\}\\---\\``Premises'': ``\{\{this.premises\}\}''\\``Proposition'': ``\{\{this.proposition\}\}''\\\{\{\/user\}\}\\\{\{\#assistant\}\}``Judgement'': ``Is this deduction valid? \{\{this.validation\}\}''\{\{\/assistant\}\}\\\{\{\~\/each\}\}
\end{mdframed}

\newpage
\begin{mdframed}[linewidth=1pt,linecolor=black]
\textbf{Prompts used for Boundary Validation}\\
\{\{\#system\}\}Suppose you are one of the greatest AI scientists, logicians, and mathematicians. Let's think about it step by step.\\
Answer ``True'' or ``False'' to determine whether the existing premises plus a new premise satisfies the boundary condition.\\
---\\
\{\{\/system\}\}\\

\{\{\~\#each examples\}\}\\
\{\{\#user\}\}\\
---\\
``existing premises'': ``\{\{this.premises\}\}''\\
``new premise'': ``\{\{this.new premise\}\}''\\
``boundary condition'': ``\{\{this.boundary condition\}\}''\\
After adding the new premise to the existing premise, does it still meet the boundary conditions?\\
\{\{\/user\}\}\\

\{\{\#assistant\}\}``Judgement'': ``\{\{this.judgement\}\}''\{\{\/assistant\}\}\\
\{\{\~\/each\}\}\\

\{\{\#user\}\}\\
---\\
``existing premises'': ``\{\{premises\}\}''\\
``new premise'': ``\{\{proposition\}\}''\\
``boundary condition'': ``\{\{boundary condition\}\}''\\
After adding the new premise to the existing premise, does it still meet the boundary conditions?\\
\{\{\/user\}\}\\

\{\{\#assistant\}\}``Judgement'': ``\{\{\/assistant\}\}\\
\{\{\#assistant\}\}\{\{select ''judgement`` options=valid duplicated\}\}\{\{\/assistant\}\}\\
\end{mdframed}

\newpage
\begin{mdframed}[linewidth=1pt,linecolor=black]
\textbf{Prompts used for Premise Transformation}\\
\{\{\#system\}\}
Suppose you are one of the greatest AI scientists, logicians, and mathematicians. Let's think about it step by step.\\
First, please read and analyze the ``existing premises'', read the definition of transformation;\\
Transformation: In the one-to-one relationship, when the value of the current variable is determined, it means that this variable can not take other values, and other variables can not take the current value, this reasoning process is transformation.\\
Check whether relying on a single ``premise'' and ``boundary condition'' can translate into other new premises? The new premises should not duplicate any of the existing premises.\\
If it can be transformed, give the new premises you have deduced; if it can't, answer ``None.''\\
Make sure that the new premises you get are helpful in solving the problem.\\
---\\
\{\{\/system\}\}\\
\{\{\~\#each examples\}\}\\
\{\{\#user\}\}\\
---\\
``existing premises'': ``\{\{this.premises\}\}''\\
``question'': ``\{\{this.question\}\}''\\
``premise'': ``\{\{this.premise\}\}''\\
``boundary condition'': ``\{\{this.boundary condition\}\}''\\
\{\{\/user\}\}\\
\{\{\#assistant\}\}Can you derive a new premise based on the premises and boundary condition that help solve the problem?\{\{\/assistant\}\}\\
\{\{\#assistant\}\}``new premise'': ``\{\{this.new premise\}\}''\{\{\/assistant\}\}\\
\{\{\~\/each\}\}\\

\{\{\#user\}\}\\
---\\
``existing premises'': ``\{\{premises\}\}''\\
``question'': ``\{\{question\}\}''\\
``premise'': ``\{\{premise\}\}''\\
``boundary condition'': ``\{\{boundary condition\}\}''\\
\{\{\/user\}\}\\
\{\{\#assistant\}\}Can you derive a new premise based on the premises and boundary condition that help solve the problem?\{\{\/assistant\}\}\\
\{\{\#assistant\}\}``new premise'': ``\{\{\/assistant\}\}\\
\{\{\#assistant\}\}\{\{gen ''premise`` temperature=temperature max tokens=50 stop=[' ']\}\}\{\{\/assistant\}\}\\
\end{mdframed}

\newpage
\begin{mdframed}[linewidth=1pt,linecolor=black]
\textbf{Prompts used for Premise Extraction}\\
\{\{\#system\}\}
Suppose you are one of the greatest AI scientists, logicians, and mathematicians. Let's think about it step by step.\\
First read and analyze the two sets of definitions defined below;\\
Premise: A constraint on the absolute position of an object or on the relative relationship between two objects.\\
Boundary condition: A description of the number of objects and the name of the object.\\
According to the above definition, summarize the core topics discussed in the following paragraphs and extract the premise and boundary conditions in the context.\\
---\\
\{\{\/system\}\}\\

\{\{\~\#each examples\}\}\\
\{\{\#user\}\}\\
---\\
``context'': ``\{\{this.context\}\}''\\
\{\{\/user\}\}\\
\{\{\#assistant\}\}Can you summarize the core topics of the discussion from the context above?\{\{\/assistant\}\}\\
\{\{\#assistant\}\}``topic'': ``\{\{this.topic\}\}''\{\{\/assistant\}\}\\
\{\{\#assistant\}\}Can you extract the premise from the context above?\{\{\/assistant\}\}\\
\{\{\#assistant\}\}``premise'': ``\{\{this.premise\}\}''\{\{\/assistant\}\}\\
\{\{\#assistant\}\}Can you extract the boundary conditions from the context above?\{\{\/assistant\}\}\\
\{\{\#assistant\}\}``boundary condition'': ``\{\{this.boundary condition\}\}''\{\{\/assistant\}\}\\
\{\{\~\/each\}\}\\

\{\{\#user\}\}\\
---\\
``context'': ``\{\{context\}\}''\\
\{\{\/user\}\}\\

\{\{\#assistant\}\}Can you summarize the core topics of the discussion from the context above?\{\{\/assistant\}\}\\
\{\{\#assistant\}\}``topic'': ``\{\{\/assistant\}\}\\
\{\{\#assistant\}\}\{\{gen ''topic`` temperature=temperature max tokens=50 stop=' '\}\}\{\{\/assistant\}\}\\
\{\{\#assistant\}\}Can you extract the premise from the context above?\{\{\/assistant\}\}\\
\{\{\#assistant\}\}''premise``: ''\{\{\/assistant\}\}\\
\{\{\#assistant\}\}\{\{gen ``premise'' temperature=temperature max tokens=300 stop=['\\n\``']\}\}\{\{\/assistant\}\}\\
\{\{\#assistant\}\}Can you extract the boundary conditions from the context above?\{\{\/assistant\}\}\\
\{\{\#assistant\}\}''boundary condition``: ''\{\{\/assistant\}\}\\
\{\{\#assistant\}\}\{\{gen ``boundary condition'' temperature=temperature max tokens=300 stop=[' ']\}\}\{\{\/assistant\}\}\\
\end{mdframed}

\newpage
\begin{mdframed}[linewidth=1pt,linecolor=black]
\textbf{Prompts used for Final Conclusion}\\
\{\{\#system\}\}Suppose you are one of the greatest AI scientists, logicians, and mathematicians. Let's think about it step by step.\\
First read and analyze the ``paragraphs'' and ``questions'', then use the ``premises'', ``boundary conditions'' and ``propositions'' to reason which of the options given is the answer to the ``question''.\\
Make sure that your reasoning is derived directly from ``premises'' and ``propositions'' rather than introducing passive common sense and passive information through common sense reasoning.\\
Please note that this is a single choice question.\\
If you can get the answer directly from the proposition, then you should choose the answer directly, otherwise keep reasoning with the proposition, premises, and boundary conditions until you arrive at a single answer.\\
---\\
\{\{\/system\}\}\\
\{\{\~\#each examples\}\}\\
\{\{\#user\}\}\\
---\\
``context'': ``\{\{context\}\}''\\
``question and options'': ``\{\{question\}\}''\\
\{\{\/user\}\}\\
\{\{\#assistant\}\}``Premises'': ``Let's think step by step, and from the context we can extract these premises: \{\{premises\}\}''\{\{\/assistant\}\}\\
\{\{\#assistant\}\}``Boundary condition'': ``Let's think step by step, and from the context we can extract these boundary conditions: \{\{boundary condition\}\}''\{\{\/assistant\}\}\\
\{\{\#assistant\}\}``Thoughts'': ``Let us think step by step. From the premises, we can deduce propositions:\{\{propositions\}\}''\{\{\/assistant\}\}\\
\{\{\#assistant\}\}``Recall the reasoning history'':``\{\{infer history\}\}''\{\{\/assistant\}\}\\
\{\{\#assistant\}\}``Recall the questions and options'':``\{\{question\}\}''\{\{\/assistant\}\}\\
\{\{\#assistant\}\}``Reasoning'': ``Using premises, boundary conditions, and continuing to reason according to the propositions already obtained,\{\{\/assistant\}\}\\
\{\{\#assistant\}\}\{\{gen ''reasoning`` temperature=0.7 max tokens=500 stop=[' ']\}\}\{\{\/assistant\}\}\\
\{\{\#assistant\}\}''Recall the questions and options``:''\{\{question\}\}``\{\{\/assistant\}\}\\
\{\{\#assistant\}\}''Judgement``: ''Now we know that the answer to this question should be\{\{\/assistant\}\}\\
\{\{\#assistant\}\}\{\{select ``judgement'' options=choose\}\}\{\{\/assistant\}\}\\
\end{mdframed}

\newpage
\section{Scalability Analysis}
Following the previous prompting methods~\cite{tot,cr} for enhanced LLM-based reasoning, our main evaluation is based on the test set (typically compact with small size) for each task. 
To further explore the scalability of the proposed method on larger and more diverse datasets, we randomly sample 1600 cases from the ProofWriter~\cite{tafjord2020proofwriter} training set to study the model performance across varying dataset sizes. The comparison results of Table~\ref{tab:scale} show that DetermLR consistently outperforms CR~\cite{cr} in terms of both accuracy and time-efficiency as data scales, which verifies the scalability and robustness of the proposed method.

\begin{table}[ht]
    \centering
    \caption{Comparison results of varying data sizes.}
    \resizebox{\textwidth}{!}{
    \begin{tabular}{lccccccc}
    \toprule
         \textbf{Data Size} & \textbf{400} & \textbf{600} & \textbf{800} & \textbf{1000} & \textbf{1200} & \textbf{1400} & \textbf{1600} \\ 
    \midrule
          Avg. Accuracy (CR) & 0.720 & 0.712 & 0.703 & 0.709 & 0.708 & 0.705 & 0.709 \\
          \textbf{Avg. Accuracy (DetermLR)} & \textbf{0.743} & \textbf{0.738} & \textbf{0.729} & \textbf{0.731} & \textbf{0.731} & \textbf{0.729} & \textbf{0.728} \\
    \midrule
Avg. Inference Time Per Case (CR) & 83.55s & 81.50s & 81.15s & 79.92s & 82.40s & 82.24s & 82.91s \\
\textbf{Avg. Inference Time Per Case (DetermLR)} & \textbf{76.20s} & \textbf{72.20s} & \textbf{68.55s} & \textbf{68.16s} & \textbf{69.05s} & \textbf{69.73s} & \textbf{70.12s} \\
\bottomrule
    \end{tabular}
    }
    \label{tab:scale}
\end{table}

\section{Generalization in Math Reasoning} 
In this paper, our primary focus is to enhance the logical reasoning capabilities of LLMs, that is, to study how LLMs better utilize the given conditions to complete the reasoning process more accurately and efficiently.
In principle, DetermLR can be easily adapted to other types of reasoning tasks---
as long as a reasoning task has a set of available conditions, the proposed method can leverage the relationship between conditions and the target to identify whether each premise is indeterminate or determinate.

Therefore, we attempt to evaluate the performance of DetermLR on other reasoning tasks, such as math reasoning. We construct a dataset of multi-variable equations. An example is shown as follows.
\begin{align}
    4x + y + 3z = 38 \quad & (1)\nonumber \\
    -x + y + z = 13 \quad & (2)\nonumber \\
    3x + 3y + z = 25 \quad & (3)\nonumber
\end{align}

Based on our insights, each equation contains quantitative relationships between multiple variables, which should be regarded as \textit{indeterminate equations}. Conversely, the exact variable values, like $x=1$, are categorized as \textit{determinate equations}. LLMs need to perform reasoning by selecting equations to combine and eliminate some variables (in line with \textit{from indeterminacy to determinacy}), and finally obtain the solution of each variable to form the overall solution.

For this evaluation, we choose CoT-SC~\cite{wang2022self} and CR~\cite{cr} as strong baselines for comparison. The accuracy and average reasoning steps per case are provided below, indicating the effectiveness and efficiency of our method.
The comparison results of the reasoning processes of DetermLR and CR are shown as follows.

We can see that the DetermLR's reasoning trajectory mirrors human-like variable elimination methods. By employing \textit{prioritization functions}, equations with fewer variables or those less frequently chosen are given precedence. Coupled with the encouragement to \textit{reduce the number of variables (from indeterminacy to determinacy)} during new equation generation, the entire reasoning process is streamlined and effective under the correct direction, and the new equations generated are useful without unnecessary redundancy. 
In contrast, CR \textit{randomly} selects these equations for combination in each round and often fails to ascertain the correct solution direction, leading to erroneous or incomplete solutions.

\newpage
\begin{mdframed}[linewidth=1pt,linecolor=black]
\textbf{Reasoning Process of CR}
\begin{enumerate}
\item In the NO:1 round,

Selected Equations: 4x + y + 3z = 38 (1), -x + y + z = 13 (2).

New Equation: 5x + 2z = 25 (4).

\item In the NO:2 round,

Selected Equations: -x + y + z = 13 (2), 3x +3y + z = 25 (3)

New Equation: 6y + 4z = 64 (5).

\item In the NO:3 round,

Selected Equations:  5x + 2z = 25 (4), 6y + 4z = 64 (5).

New Equation: 6y - 10x = 14 (6).

\item In the NO:4 round,

Selected Equations: 5x + 2z = 25 (4), 6y - 10x = 14 (6).

New Equation: 6y + 4z = 64 (7)

\item In the NO:5 round,

Selected Equations: -x + y + z = 13 (2), 5x + 2z = 25 (4).

New Equation: -7x + 2y = 1 (8).

\item In the NO:6 round,

Selected Equations: 3x +3y + z = 25 (3), 5x + 2z = 25 (4).

New Equation: x + 6y =25 (8).

\item ...
\end{enumerate}
\textbf{Solution:} It's clear that the system is becoming more complicated ... (\textcolor{red}{Wrong})
\end{mdframed}

\newpage
\begin{mdframed}[linewidth=1pt,linecolor=black]
\textbf{Reasoning Process of DetermLR}
\begin{enumerate}
\item In the NO:1 round,

Selected Equations: 4x + y + 3z = 38 (1), -x + y + z = 13 (2).

New Equation: 5x + 2z = 25 (4).

\item In the NO:2 round,

Selected Equations: 5x + 2z = 25 (4), 3x +3y + z = 25 (3)

New Equation: x + 6y = 25 (5).

\item In the NO:3 round,

Selected Equations:  -x + y + z = 13 (2), 5x + 2z = 25 (4).

New Equation: -7x + 2y = 1 (6).

\item In the NO:4 round,

Selected Equations: x + 6y = 25 (5), -7x + 2y = 1 (6).

New Equation: x = 1 (7)

\item In the NO:5 round,

Selected Equations: x = 1 (7), x + 6y = 25 (5).

New Equation: y = 4 (8).

\item In the NO:6 round,

Selected Equations: x = 1 (7), 5x + 2z = 25 (4).

New Equation: z = 10 (8).
\end{enumerate}
\textbf{Solution:} x=1, y=4, z=10. (\textcolor{green}{Right})
\end{mdframed}


\end{document}